\PassOptionsToPackage{unicode}{hyperref}
\PassOptionsToPackage{hyphens}{url}
\documentclass[sigconf]{acmart}
\usepackage[T1]{fontenc}
\usepackage[utf8]{inputenc}
\input{glyphtounicode}
\usepackage{xcolor}
\IfFileExists{upquote.sty}{\usepackage{upquote}}{}
\IfFileExists{microtype.sty}{\usepackage[]{microtype}\UseMicrotypeSet[protrusion]{basicmath}}{}
\usepackage{longtable,booktabs,array}
\usepackage{calc}
\usepackage{etoolbox}
\makeatletter
\patchcmd\longtable{\par}{\if@noskipsec\mbox{}\fi\par}{}{}
\makeatother
\IfFileExists{footnotehyper.sty}{\usepackage{footnotehyper}}{\usepackage{footnote}}
\makesavenoteenv{longtable}
\usepackage{graphicx}
\usepackage{float}
\usepackage[font=footnotesize,labelfont=bf]{caption}
\usepackage{amsmath}
\usepackage{tikz}
\usetikzlibrary{arrows.meta,calc,positioning,fit}
\usepackage{enumitem}
\makeatletter
\newsavebox\pandoc@box
\newcommand*\pandocbounded[1]{%
  \sbox\pandoc@box{#1}%
  \Gscale@div\@tempa{\textheight}{\dimexpr\ht\pandoc@box+\dp\pandoc@box\relax}%
  \Gscale@div\@tempb{\linewidth}{\wd\pandoc@box}%
  \ifdim\@tempb\p@<\@tempa\p@\let\@tempa\@tempb\fi%
  \ifdim\@tempa\p@<\p@\scalebox{\@tempa}{\usebox\pandoc@box}%
  \else\usebox{\pandoc@box}%
  \fi%
}
\def\fps@figure{htbp}
\makeatother
\providecommand{\tightlist}{\setlength{\itemsep}{0pt}\setlength{\parskip}{0pt}}
\usepackage{bookmark}
\IfFileExists{xurl.sty}{\usepackage{xurl}}{}
\hypersetup{hidelinks,pdfcreator={LaTeX via pandoc}}

\title{Causal Learning with Neural Assemblies}

\author{Evangelia Kopadi}
\affiliation{%
  \institution{Hellenic Open University}
  \country{Greece}}

\author{Dimitris Kalles}
\affiliation{%
  \institution{Hellenic Open University}
  \country{Greece}}

\acmConference{}{}{}
\acmBooktitle{}
\acmYear{2026}
\acmDOI{}
\acmISBN{}
\setcopyright{cc}\setcctype{by-nc-nd}

\begin{document}
\raggedbottom
\maketitle

\section*{Abstract}\label{abstract}

Can Neural Assemblies---groups of neurons that fire together and strengthen through co-activation---learn the direction of causal influence between variables? While established as a computationally general substrate for classification, parsing, and planning, neural assemblies have not yet been shown to internalize causal directionality.

We demonstrate that the inherent operations of neural assemblies---projection, local plasticity control, and sparse winner selection---are sufficient for directional learning. We introduce DIRECT (DIRectional Edge Coupling/Training), a mechanism that co-activates source and target assemblies under an adaptive gain schedule to internalize directed relations. Unlike backpropagation-based methods, DIRECT relies solely on local plasticity, making the resulting causal claims auditable at the mechanism level.

Our findings are verified through a dual-readout validation strategy: (i) synaptic-strength asymmetry ($\Delta$), measuring the emergent weight gap between forward and reverse links, and (ii) functional propagation overlap ($\Delta_{\text{prop}}$), quantifying the reliability of directional signal flow. Across multiple domains, the framework achieves perfect structural recovery (Precision@K = 1.0) under a supervised, known-structure setting. These results establish neural assemblies as an auditable bridge between biologically plausible dynamics and formal causal models, offering an "explainable by design" framework where causal claims are traceable to specific neural winners and synaptic asymmetries.

\section{Introduction}\label{introduction}

Our motivation is to investigate causal direction learning through a mechanism that is both locally learnable and directly inspectable. Unlike opaque global optimizers or backpropagation, Neural Assemblies learn through local plasticity alone, exposing their learned structure through identifiable winners and measurable synaptic asymmetry \cite{hebb1949,buzsaki2010,markram1997,bi1998,song2000}. This approach makes causal claims auditable at the mechanism level while remaining grounded in biologically plausible neural computation.

To provide a clear conceptual framework, we begin with structural requirements that include the definition of variables, their value-range mapping, and the ground-truth causal relations. On that foundation, representation transforms categorical data (variables recorded as discrete labels, such as low/medium/high or yes/no) into sparse neural spike patterns, which become the fundamental language of the brain simulator and make the inputs compatible with biologically plausible neural dynamics. The process then moves to stabilization and directed binding, where encoded patterns converge toward stable, reusable Neural Assemblies through recurrent co-activation while directed links are strengthened with DIRECT under an adaptive warm-ramp gain progression. This keeps directional updates conservative during early winner-set drift (instability in which neurons are selected as winners across repeated presentations of the same pattern) and increases binding strength as stability evidence accumulates, preserving asymmetric forward signal flow. Finally, auditability is established by verifying causal claims through synaptic-weight asymmetry and functional propagation readout, so the full mechanism remains explainable by design.

\begin{figure}[H]
\centering
\includegraphics[width=0.92\linewidth]{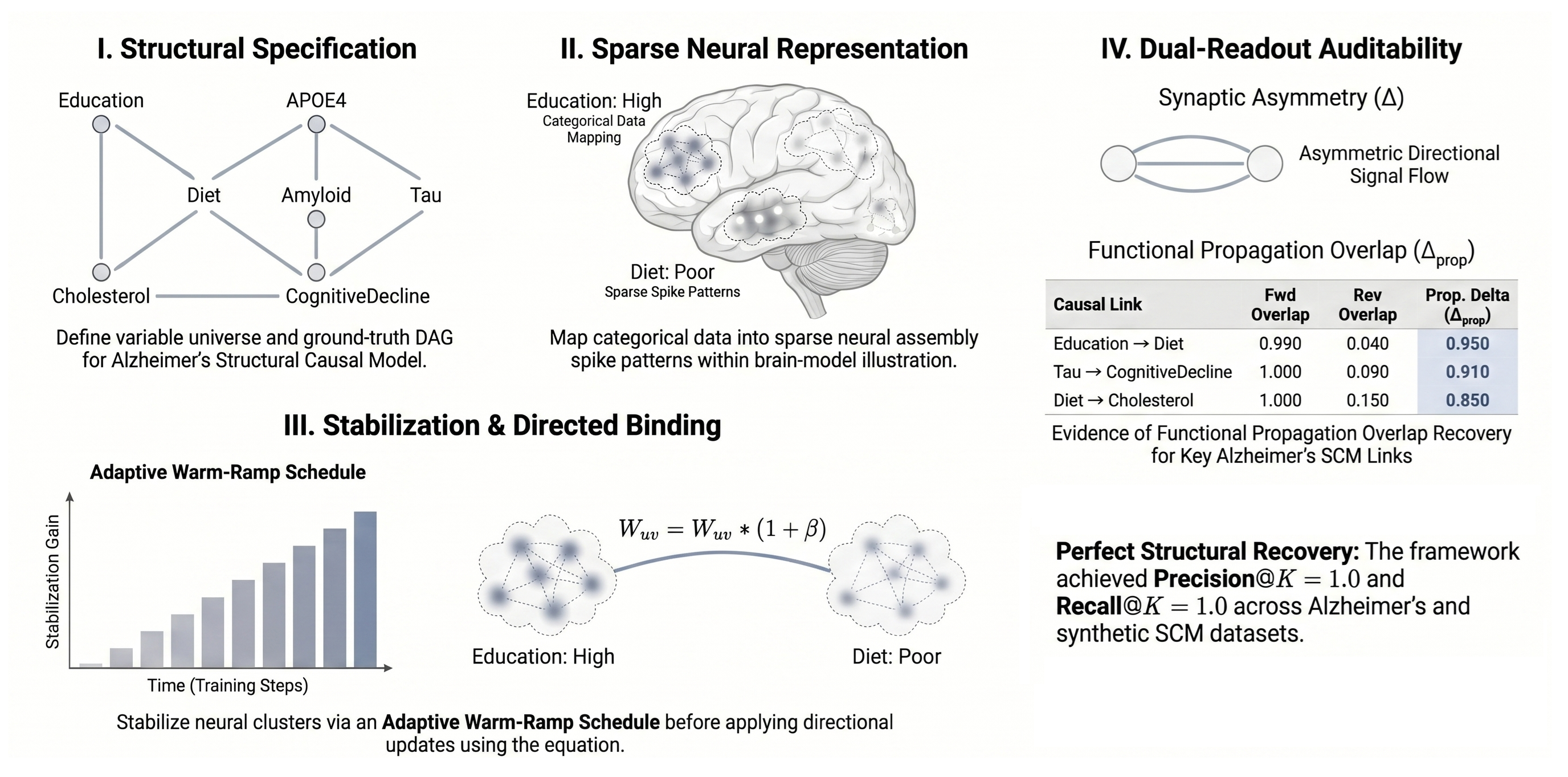}
\caption{Pipeline overview.}\label{fig:pipeline}
\vspace{-0.45\baselineskip}
\end{figure}

This investigation continues the same research trajectory developed in section 7. In the first phase, we established that neural assemblies can preserve causal information across a full end-to-end process, from raw variables to neuron and assembly feature spaces, under controlled conditions.

Building on that foundation, the present phase asks the next logical question: can assemblies go beyond preserving existing structure and actively learn causal directionality between variables? This step is critical for establishing neural assemblies as a computationally general substrate for structured cognition. Prior work in the neural-assemblies line supports this broader computational framing \cite{papadimitriou2020,mitropolsky2021,dabagia2025}. Causal reasoning is fundamental to human intelligence and decision-making within formal causal semantics \cite{spirtes2000,pearl2009,peters2017}. By showing that the same operations that form stable representations can also internalize directed structure, we position neural assemblies not only as passive encoders of information, but as active learners of causal relationships. 

In this work, we do not claim de novo causal discovery from observational data alone; instead, we evaluate whether the mechanism internalizes provided directional evidence. Specifically, we investigate whether neural assemblies can internalize directed structure in a way that supports auditable causal claims under a deliberately bounded setting. The DAG/SCM structure is provided as ground truth, and the task is restricted to supervised learning of directional bindings over that fixed structure. By establishing representational stability before directional binding, we attribute directional effects to the DIRECT mechanism rather than representational drift, enabling stage-level diagnostic localization. We therefore position the framework as a biologically plausible mechanism-class abstraction compatible with Hebbian/STDP-style plasticity, not as a circuit-faithful biological model.

\section{Related Work}\label{related-work}

This section defines the framework that links neural-assembly-level dynamics to causal claims. 

The framework is grounded in three theoretical pillars. First, neural assembly representation theory: cell assemblies from Hebb and their modern formalization by Papadimitriou and collaborators, including NEMO-line results by Dabagia, Papadimitriou, Vempala, and Mitropolsky \cite{hebb1949,buzsaki2010,papadimitriou2020}. Second, directional plasticity: order-asymmetric synaptic learning from core spike-timing-dependent plasticity (STDP) literature \cite{markram1997,bi1998,song2000}. Third, causal semantics: DAG/SCM-based interpretation from the causal-graph tradition and Pearl's interventional do-calculus \cite{spirtes2000,pearl2009,peters2017}.

The representation process (neural assembly theory \cite{hebb1949,buzsaki2010,papadimitriou2020}): repeated co-activation can produce sparse, stable attractor-like states (activity patterns that reliably settle into the same configuration) that serve as reusable variable-level representations. In the NEMO research line (Papadimitriou, Vempala, Dabagia, Mitropolsky), this mechanism has been demonstrated across a broad range of computational settings: assembly formation and recall \cite{papadimitriou2020}, classification of well-separated distributions \cite{papadimitriou2020,dabagia2022}, biologically plausible syntactic parsing \cite{mitropolsky2021}, planning \cite{damore2022}, computation with sequences of assemblies including finite-state machine simulation \cite{dabagia2025}, and statistical learning of probability distributions \cite{dabagia2024}. Taken together, these results suggest that neural assemblies are approaching computational generality as a substrate. The present work extends this trajectory by showing that the same operations can also internalize causal directionality. Under this principle, directional inference is meaningful only after representational states are sufficiently stable; otherwise, apparent link direction may be confounded by representational drift.

The causal directionality process (asymmetric plasticity \cite{markram1997,bi1998,song2000}): when learning dynamics are order-asymmetric, expected forward and reverse couplings can systematically diverge. This yields an observable directional signal at mechanism level (forward-minus-reverse asymmetry) that is interpretable without requiring full biophysical realism.

The semantics (SCM \cite{spirtes2000,pearl2009,peters2017}): causal claims are valid only relative to structural assumptions. We follow the DAG/SCM lineage from graph-based causal modeling to modern intervention semantics, including Pearl's do-calculus (rules for reasoning about interventions of the form do(X=x)). Because the DAG is externally provided, the estimand in this work is supervised mechanism-learning quality conditional on ground truth (the specific quantity we aim to estimate), not de novo graph discovery from observational data.

Taken together, these three main theories constitute the basis for the paper's contribution: a model linking neural-assembly-level dynamics to causal-mechanism evidence.

\section{Neural Assemblies Architecture To Learn Causality}\label{neural-assemblies-architecture-to-learn-causality}

Three design commitments follow from the theoretical principles above. First, we use a single adaptive schedule rather than a hard stage boundary: directional updates begin with low gain during a stabilization-aware warmup, then smoothly ramp to the target binding gain after overlap-based convergence signals are met. This adaptive warm-ramp design preserves representational stability constraints without requiring a hard freeze, and keeps directional asymmetry attributable to bounded gain modulation over already-stabilizing winner sets. The selected setting (\texttt{adaptive\_soft}, \texttt{warm\_beta = 0.09}, \texttt{ramp\_steps = 20}) was retained after preliminary tuning because it gave the best balance between early representational stability and later directional separation. A conservative warm gain (warm\_beta = 0.09) limits winner-set drift during initial rounds, while the 20-step ramp increases binding strength gradually rather than abruptly. In the reported runs, this schedule consistently preserved forward-minus-reverse asymmetry and stable Top-K recovery.

\begin{figure}[H]
\centering
\includegraphics[width=0.92\linewidth]{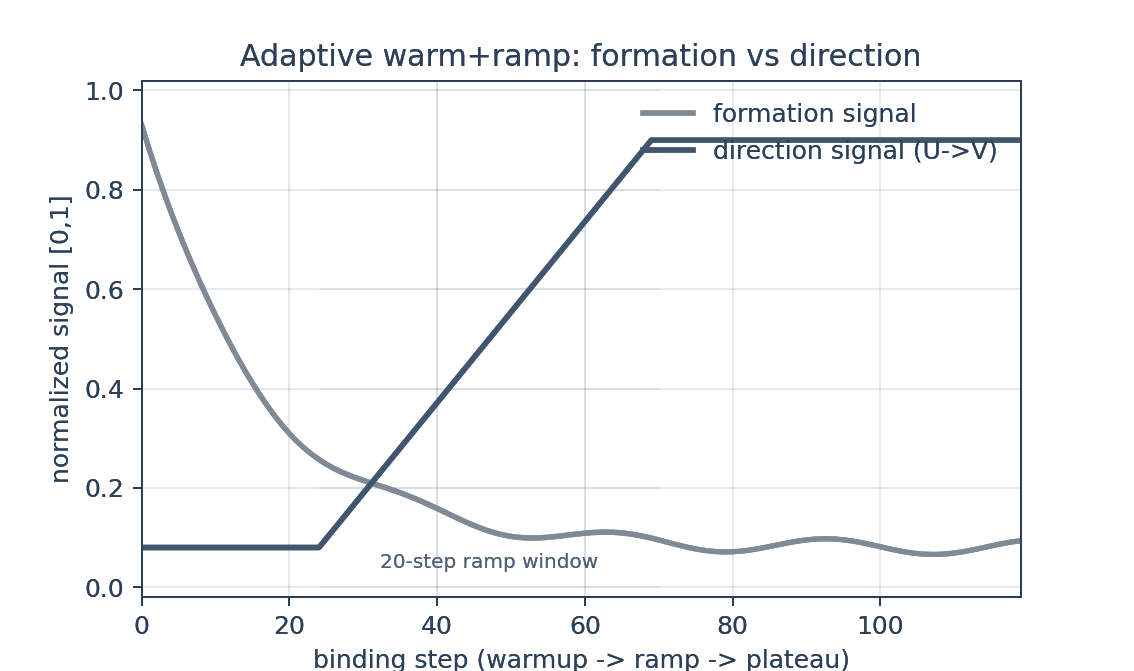}
\caption{Warm+ramp success diagnostic.}
\end{figure}

The need for this adaptive schedule becomes apparent when compared with the baseline parallel setting, where formation and directional updates run concurrently without adaptive tuning. In that regime, winner drift remains high, directional signal becomes noisy, and edge-level asymmetry is harder to attribute mechanistically.

\begin{figure}[H]
\centering
\includegraphics[width=0.92\linewidth]{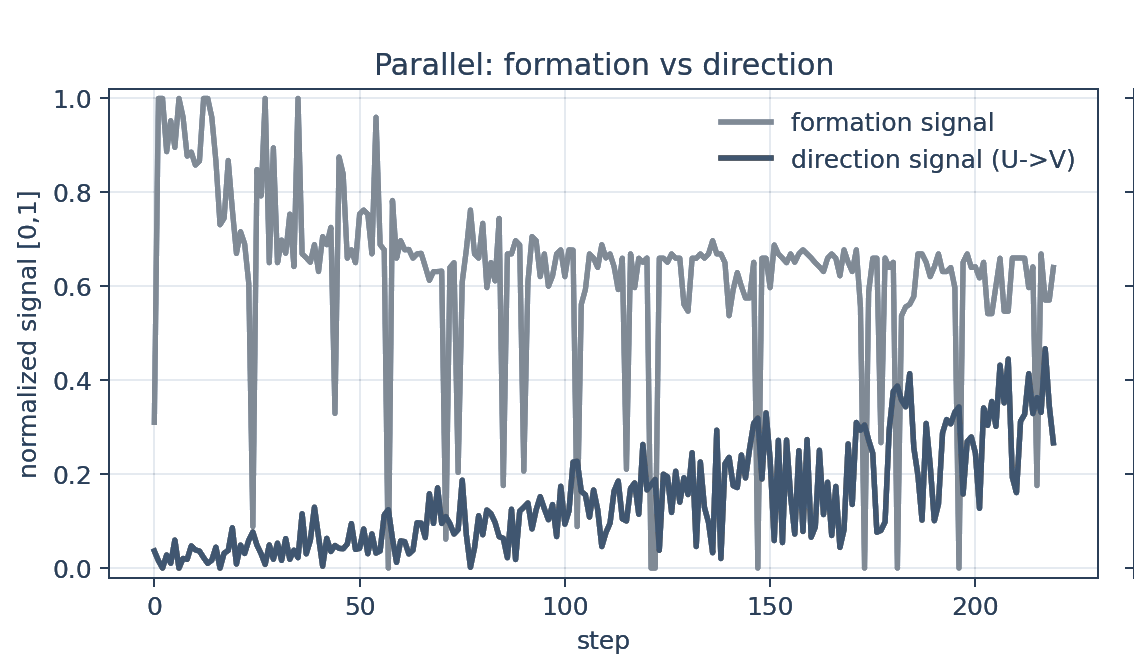}
\caption{Parallel failure diagnostic.}
\end{figure}

Second, a single Brain simulator is shared across all stages, ensuring that formation and binding operate on the same connectome (the model’s full synaptic wiring matrix) and plasticity state. 

Third, each stage and the subsequent readout step are independently inspectable, so that if a directed link is not recovered, the discrepancy can be localized to assembly instability, insufficient binding asymmetry, or readout evaluation. This stage-level auditability is a key strength of the framework, allowing for mechanism-level failure analysis and interpretability.

\section{Causal Information Preservation via Neural Assemblies}\label{causal-information-preservation-via-neural-assemblies}

First, we tested whether neural assemblies \cite{papadimitriou2020} preserve causal information across a full end-to-end process rather than only at a single analysis stage. Starting from synthetic datasets generated from known Structural Causal Models (SCMs; equation-based causal data-generating systems) and their ground-truth Directed Acyclic Graphs (DAGs; directed graphs without cycles) \cite{spirtes2000,pearl2009}, we encoded categorical variables into sparse population codes, formed assemblies through k-Winner-Take-All (k-WTA; only the top-k neurons remain active) competition with Hebbian plasticity (co-active neurons strengthen connections) \cite{hebb1949,papadimitriou2020}, and extracted low-dimensional neuron and assembly features for downstream causal analysis. 

\begin{figure}[H]
\centering
\includegraphics[width=0.92\linewidth]{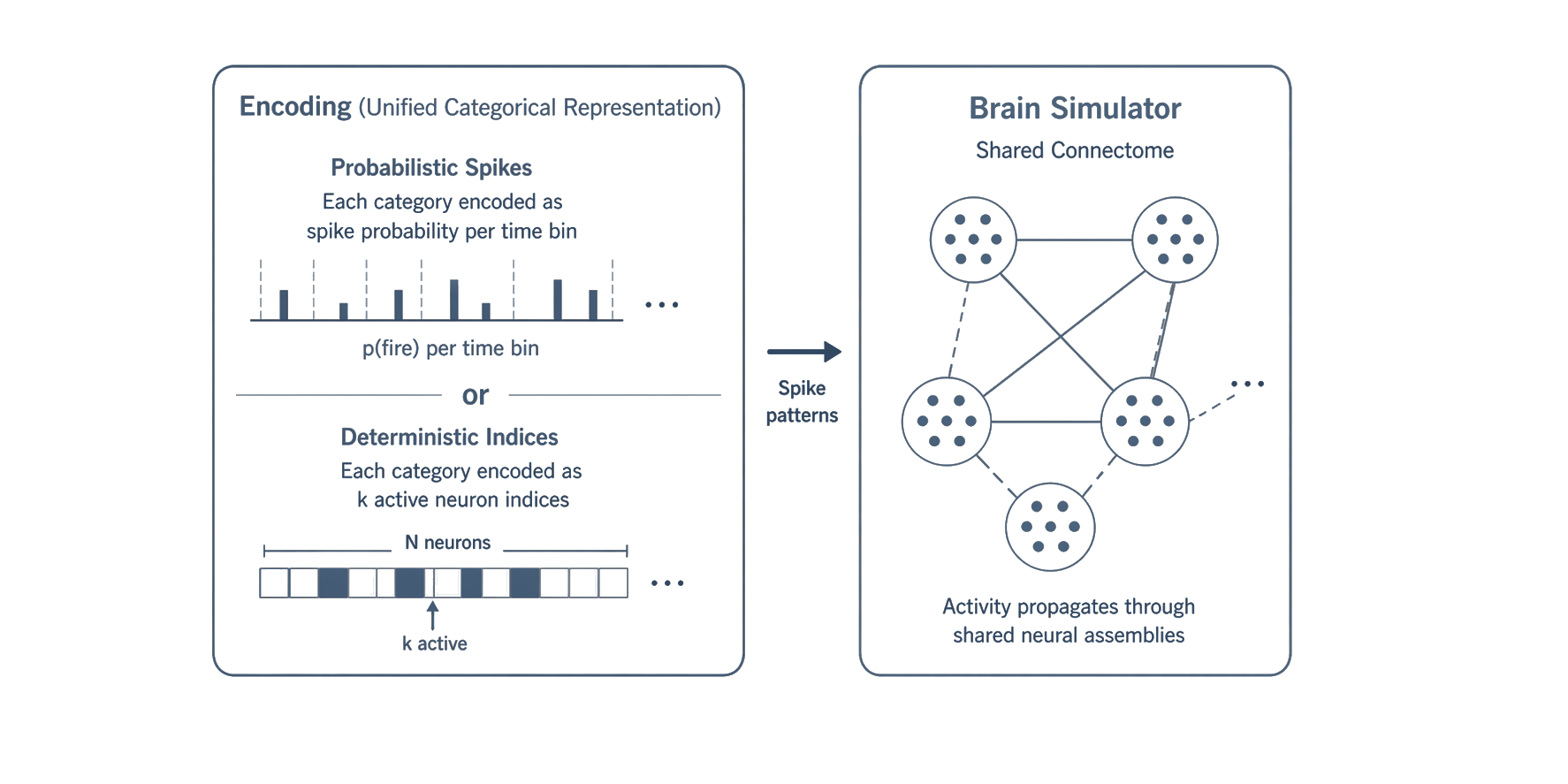}
\caption{Mixed-input encoding overview.}
\end{figure}

Across domains, causal discovery on assembly features closely matched discovery on neuron features when using the PC algorithm (Peter-Clark conditional-independence method) \cite{spirtes2000}, and largely recovered the ground-truth structure. Interventional checks using do-operator data generation (do(X=x); explicit intervention that sets a variable value in the SCM) \cite{pearl2009} further showed that causal-effect direction and relative effect magnitude were preserved from raw variables to neuron and assembly feature spaces.

After Neural Encoding, the process bifurcates into two parallel paths: Path A extracts neuron-level scalar features and applies the PC algorithm (Peter-Clark conditional-independence method) \cite{spirtes2000} to recover a Neuron DAG, while Path B passes activity through Assembly Formation (Step 3: k-WTA + Hebbian), then extracts assembly-level scalar features and applies the same PC algorithm to recover an Assembly DAG. The comparison target is direct: whether the Assembly DAG matches, or is more robust than, the Neuron DAG relative to ground truth.

Crucially, both paths --- the raw neuron-level and the assembly-level --- yielded equivalent causal recovery results across all tested domains: the Assembly DAG matched the Neuron DAG relative to ground truth, confirming that assembly formation preserves rather than distorts causal structure. A detailed presentation of these experiments is beyond the scope of this paper; we report the conclusion here as the empirical premise for the present work. Because both encoding strategies (value-to-firing-rate and NEMO-like index-based) produced equivalent top-line DAG recovery in the preservation phase, either technique may be used in the section \ref{feature-encoding} described below.

A compact visualization of this seven-step process is shown below.

\begin{figure}[H]
\centering
\includegraphics[width=0.92\linewidth]{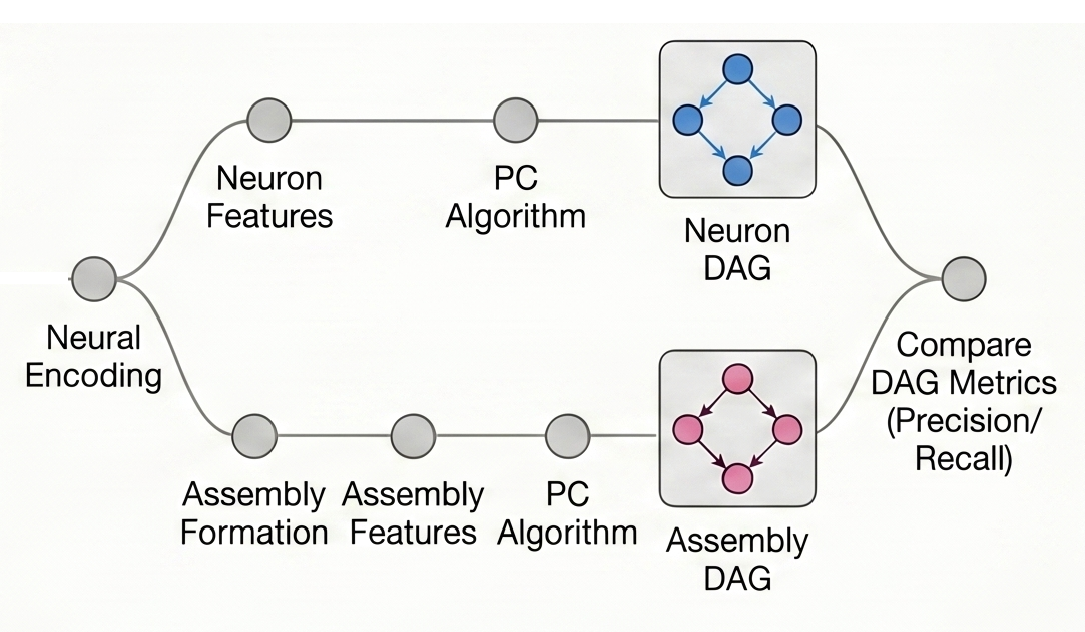}
\caption{Causal information preservation process.}
\end{figure}

\section{Causal Learning Methodology}\label{causal-learning-methodology}

The pipeline is domain-agnostic. This section specifies the stage logic independently of dataset identity. A domain dataset provides three items only: ordered variables, directed links (ground truth) and positive value mapping. The pipeline core remains unchanged across domains.

We encode training rows, form neural assemblies, apply directed causal binding per link under temporary forward plasticity gain, restore baseline plasticity controls, compute ordered-pair asymmetry scores, extract thresholded Top-K directed hypotheses (K fixed to ground-truth link count), then summarize structural metrics across single-run and fold-partition runs.

The pipeline overview figure 1 summarizes this stage order at a glance. Each stage has a single responsibility and a directly inspectable output, which supports mechanism-level auditability and failure analysis.

\subsection{Data Ingestion}\label{data-ingestion}

Input data follow a shared tabular format across domains. Each row is an observation, each column is a variable, and each variable is encoded into a value category used by the defined positive value mapping. This separation keeps dataset-specific preprocessing outside the core causal-learning pipeline.

For time-aware variants, temporalization is handled in Data Ingestion rather than in Feature Encoding. Concretely, we construct lagged/current variable pairs and constrain admissible supervision links to forward-in-time edges before neural encoding is applied. In the present manuscript, this is an exploratory alternative training setup with limited experiments, not a main validated path. In practice, reliable direction recovery in this setting still depends on additional directional binding under plasticity asymmetry (DIRECT-style forward bias), so we do not treat temporal indexing alone as sufficient evidence of causal direction learning.

From a theoretical perspective, this stage is defined by four structural requirements and is independent of file syntax: (1) a variable universe with stable ordering, (2) an observation function that maps each variable to a categorical state per sample, (3) a polarity map that marks the intervention-relevant state for each variable, and (4) a supervision graph specifying admissible directed links. Any dataset representation that satisfies these requirements is equivalent for the learner. The three items provided by the dataset are:

\begin{itemize}
\tightlist
\item \textbf{Raw dataset}: domain-native tables/files. Role: source only; not consumed directly by the learner.
\item \textbf{Model input table}: one row per instance and one column per modeled variable. Role: consumed by Neural Assembly Formation and Directed Causal Binding.
\item \textbf{Specification JSON}: variable ordering, link list, and variable value mapping. Role: defines ordering, supervision target, and variable value range.
\end{itemize}

The same structural setup is independent of any specific dataset origin: only the mapped table and task specification are consumed by the learner.

\subsection{Feature Encoding}\label{feature-encoding}

Two encoding modes are supported as described in section \ref{causal-information-preservation-via-neural-assemblies}. Both use the same tabular input and differ only in the representation passed into brain dynamics. The same encoding procedure is applied identically to temporally indexed variables, once those variables are defined in Data Ingestion.

\begin{itemize}
\tightlist
\item \textbf{Value-to-firing-rate}: each neuron fires independently with a value-dependent probability (e.g., 0.30 for positive, 0.10 for negative polarity). The stimulus is then summarized by its total firing count. For instance, encoding a positive-polarity value across 1666 neurons produces roughly 500 active neurons per sample, with the exact count varying stochastically across presentations. Biological proximity: lower (stronger abstraction). Operational usability: higher (simpler and easier to reproduce).
\item \textbf{Index-based active set (NEMO-like) \cite{papadimitriou2020,dabagia2022}}: each unique value is mapped to a fixed set of k neuron indices, which are injected directly into the input area (NEMO-like here means identity-preserving sparse index coding rather than rate-only coding). For instance, encoding a value with k = 100 always activates the same 100 predetermined neurons (e.g., indices {3, 17, 42, ...}), making the representation fully deterministic across samples. Biological proximity: higher (identity-level sparse pattern preserved). Operational usability: lower (more wiring/debug complexity).
\end{itemize}

At current evaluated operating points, both modes show the same top-line DAG recovery, with no consistent runtime advantage. Selection is therefore primarily methodological: representation fidelity versus operational simplicity. In our earlier causal-information-preservation experiments, a constant-k variant of the NEMO-like encoding suffered from signal collapse — different values produced indistinguishable mean activations. This was resolved by using value-dependent assembly sizes (\texttt{k = base\_k + i × step}, where \texttt{i} is the value index), which achieved F1 > 0.94 for DAG recovery. The present work uses the value-to-firing-rate encoding for its operational simplicity and robustness; the NEMO-like mode remains available as an alternative.

\subsection{Assembly Formation}\label{assembly-formation}

Assembly Formation stabilizes variable-level assemblies from encoded training rows through repeated exposure under local co-activation only, with no directional gain. In line with assembly theory, repeated projection under sparse winner competition (k-WTA) and local Hebbian plasticity reinforces co-active neuron sets, increases winner overlap across rounds, and strengthens within-assembly coupling, yielding sparse reusable attractor-like representations with reduced representational drift \cite{hebb1949,buzsaki2010,papadimitriou2020}. In the adaptive warm-ramp regime, assembly formation remains an ongoing stabilization process, with winner overlap and within-assembly coupling continuing to consolidate as training proceeds; the gain schedule is defined in Directional Binding.

\begin{figure}[H]
\centering
\includegraphics[width=0.92\linewidth]{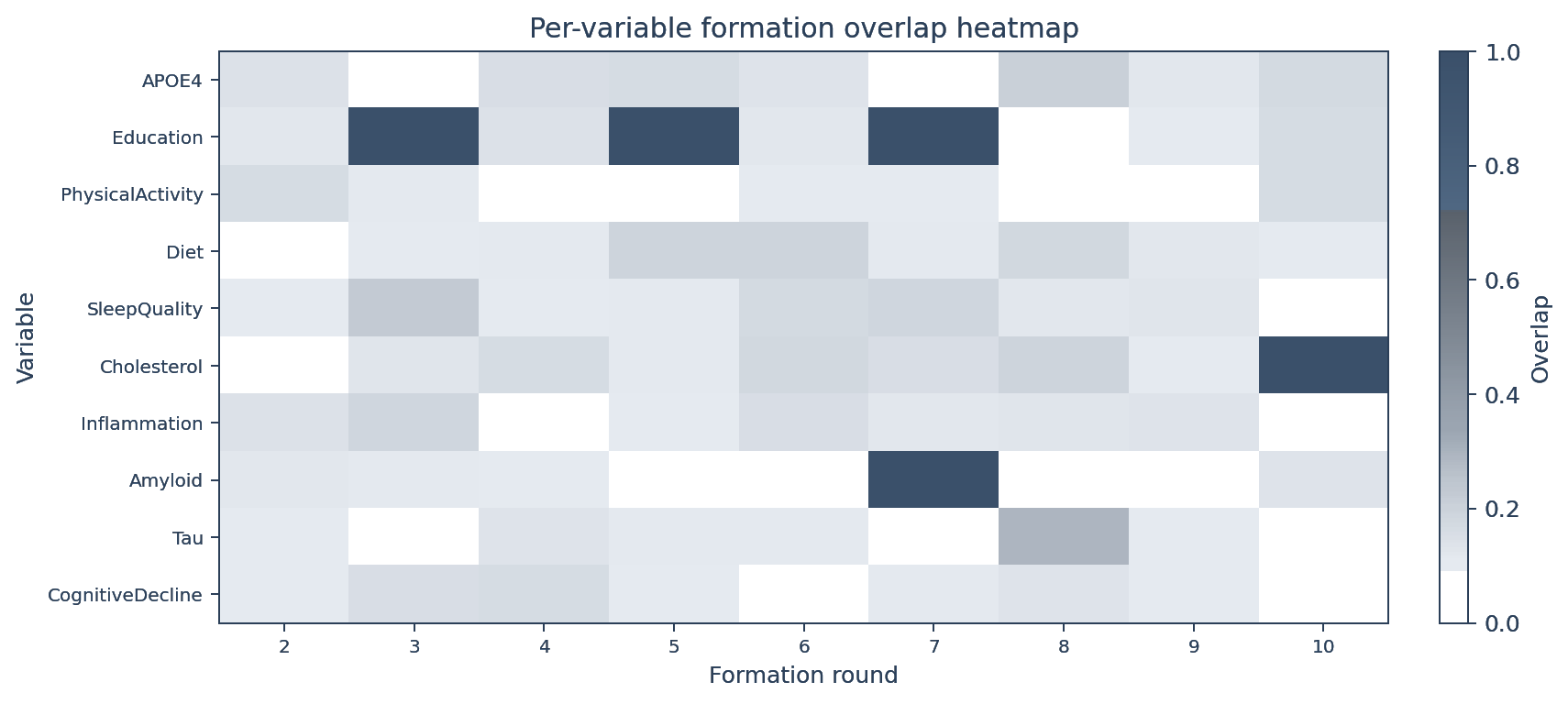}
\caption{Assembly-stability heatmap after formation rounds.}
\end{figure}

\subsection{Directional Binding}\label{directional-binding}

A central challenge in representation learning is distinguishing statistical dependence from causal structure. While correlated variables exhibit symmetric co-variation, causal relationships are inherently directional and manifest through asymmetric responses under intervention. In purely observational settings, this asymmetry is typically not directly accessible, as standard measures such as correlation or mutual information are symmetric and therefore insufficient to infer directionality.

The objective is to establish known directed links between the fixed neural assemblies layer. We do that by repeatedly presenting known link pairs under temporary forward gain modulation, while preserving the underlying neural assemblies. The output is asymmetric forward-vs-reverse activation patterns that support directed link readout.

We refer to this binding process as \textbf{DIRECT} (DIRectional Edge Coupling/Training): for each ground-truth link \texttt{(u -> v)}, the same assembly operations — projection, local plasticity control, and sparse winner selection — are applied in a specific order that produces directional learning. No new operation is introduced. 
DIRECT takes the stabilized source and target assemblies (A\_u, A\_v), applies a bounded-time increase in forward plasticity gain g\_dir > 1 during co-activation for T exposures, and then restores plasticity parameters to baseline.

Operationally, each exposure co-activates the pair while the forward pathway is temporarily gain-modulated. The implemented update is winner-sparse and multiplicative:

\begin{itemize}
\tightlist
\item the procedure temporarily increases plasticity only for the supervised forward link \texttt{(u -> v)}:
\end{itemize}

\begin{quote}\small\noindent
\texttt{update\_plasticity(}\\[0pt]
\hspace*{2em}\texttt{from\_area=u,}\\[0pt]
\hspace*{2em}\texttt{to\_area=v, new\_beta=bind\_beta)}
\end{quote}

\begin{itemize}
\tightlist
\item during projection, for each active source winner \texttt{j} and target winner \texttt{i}, the connectome entry is updated as:
\end{itemize}

\begin{quote}\small\noindent
\texttt{the\_connectome[j, i]}\\[0pt]
\hspace*{2em}\texttt{*= 1.0 + area\_to\_area\_beta}
\end{quote}

\begin{itemize}
\tightlist
\item because \texttt{area\_to\_area\_beta} is boosted only on the forward link during the binding window, forward links are strengthened more than reverse links;
\item after the binding steps, plasticity is restored to its previous value.
\end{itemize}

\begin{figure}[H]
\centering
\includegraphics[width=0.92\linewidth]{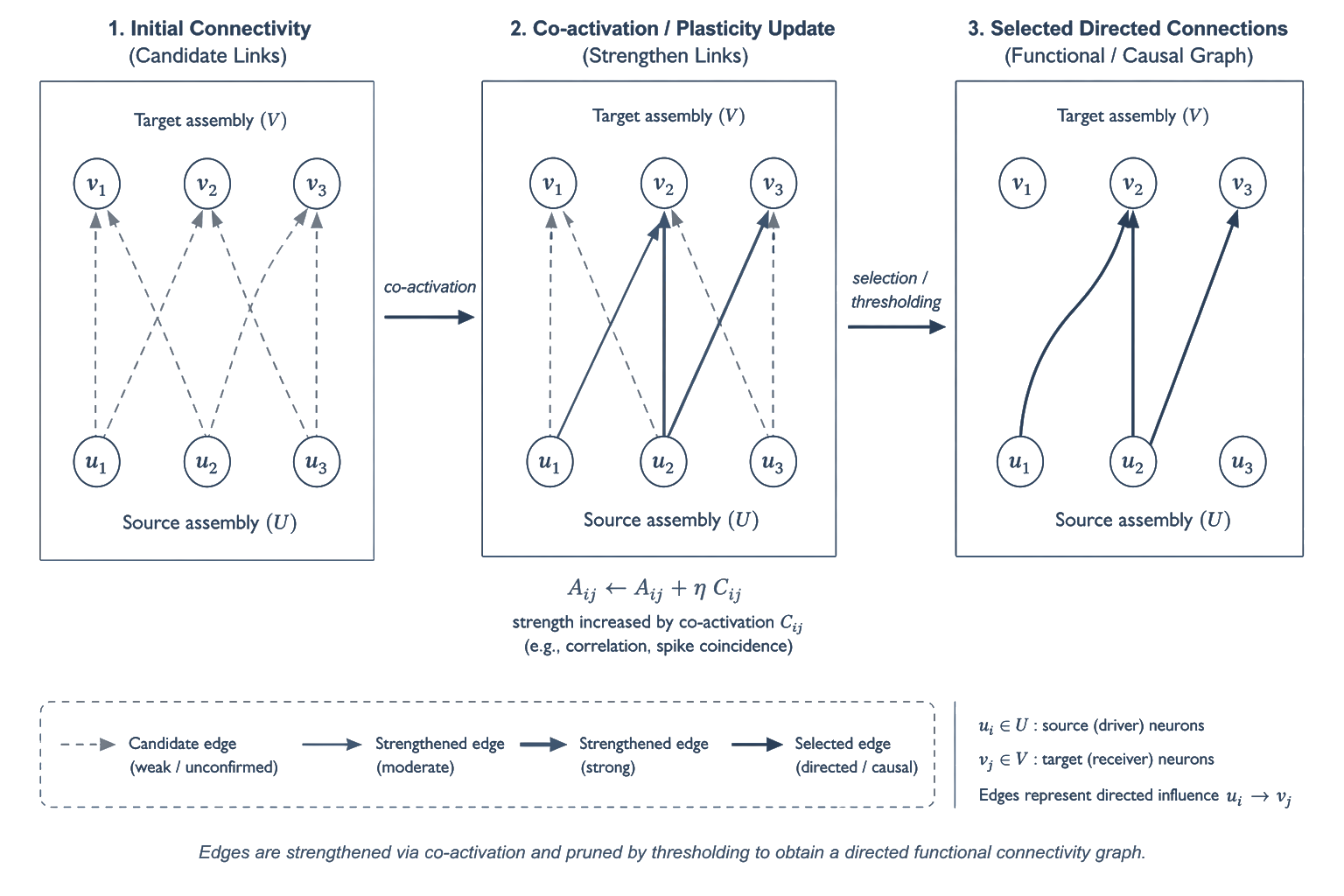}
\caption{DIRECT mechanism overview.}
\end{figure}

\emph{Clarification:} The mechanism does not remove reverse links. For every supervised pair \texttt{(u -> v)}, both \texttt{(u -> v)} and \texttt{(v -> u)} pathways exist; DIRECT increases forward plasticity temporarily, so reverse remains present but weaker. The directional signal emerges from the asymmetry in strength between forward and reverse pathways, not from the presence or absence of connections. Temporary’ refers to the duration of gain modulation, not to automatic reversal of synaptic updates.

\textbf{Example (pair input and readout).} Consider a specified pair \texttt{(u -> v)}. DIRECT repeatedly inputs the assembly pair \texttt{(A\_u, A\_v)} during Directional Binding episodes. After training, directionality is read out from the same pair using:

\begin{itemize}
\tightlist
\item Forward score: \texttt{S\_fwd = mean(A\_u -> A\_v)}
\item Reverse score: \texttt{S\_rev = mean(A\_v -> A\_u)}
\item Directional gap: \texttt{Delta(u,v) = S\_fwd - S\_rev}
\end{itemize}

A positive \texttt{Delta(u,v)} and high rank in the ordered-pair list constitute directional evidence for \texttt{u -> v} under the ground-truth setting.

\subsection{Causal Readout}\label{causal-readout}

\subsubsection{Synaptic-Strength Readout}\label{synaptic-strength-readout}

Directional evidence is computed from forward/reverse synaptic-strength summaries extracted from learned inter-area connectomes. All ordered pairs are scored and ranked, then filtered by the ratio threshold. Top-K keeps the first K ranked pairs. Here K is fixed to the number of ground-truth links, so the predicted link-set size matches the ground-truth link-set size and keeps precision/recall comparable across runs and domains.

\begin{figure}[H]
\centering
\includegraphics[width=0.92\linewidth]{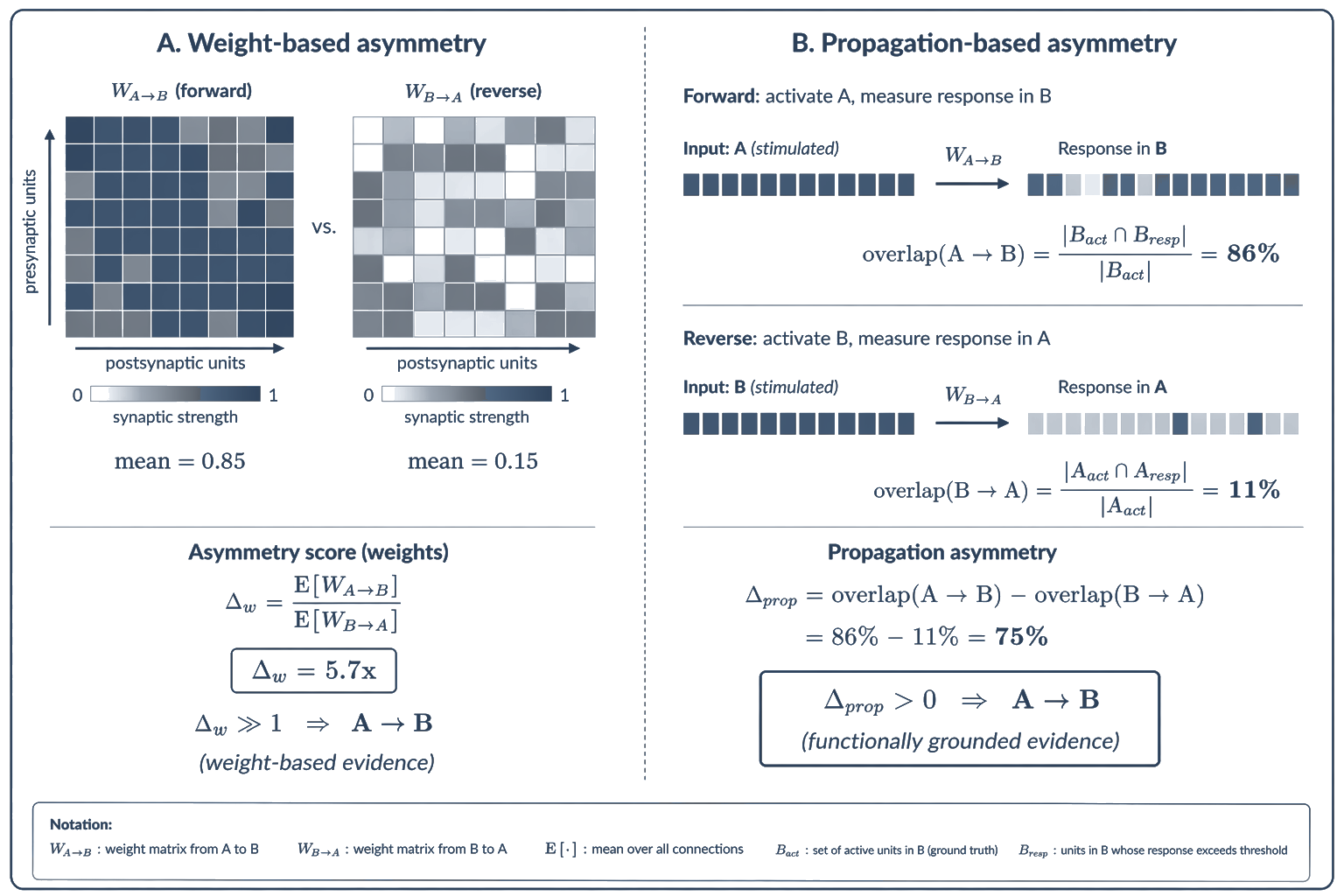}
\caption{Readout comparison summary.}
\end{figure}

The selected reporting regime is the adaptive soft schedule (\texttt{adaptive\_soft}, \texttt{warm\_beta = 0.09}, \texttt{ramp\_steps = 20}), used consistently in the Alzheimer adaptive experiments and robustness analyses.

Explicitly, this is one continuous adaptive gain schedule: directional gain starts at a conservative value (\texttt{bind\_beta = warm\_beta}) while overlap is monitored each round, and then smoothly increases once stability evidence accumulates (overlap at least \texttt{overlap\_thr} for \texttt{stable\_window} consecutive rounds, or the warmup cap if this criterion is not reached earlier). The increase is linear from \texttt{beta\_start = max(warm\_beta, 0.06)} to \texttt{max\_beta} over \texttt{ramp\_steps}, avoiding any hard jump:

\[
\beta_s = \beta_{\text{start}} + \frac{s-1}{R-1}\left(\beta_{\max}-\beta_{\text{start}}\right), \quad s=1,\dots,R, \; R=\text{ramp\_steps}.
\]

For the chosen configuration (\texttt{warm\_beta = 0.09}, \texttt{max\_beta = 0.16}, \texttt{ramp\_steps = 20}), the linear increment is approximately $\Delta\beta \approx 0.00368$ per ramp step.

\subsubsection{Propagation Overlap Readout}\label{propagation-overlap-readout}

In addition to the synaptic-strength ratio, we validate directional recovery through a functional propagation test inspired by the NEMO classification paradigm \cite{papadimitriou2020,dabagia2022}. In NEMO, a new stimulus is projected through trained weights with plasticity disabled, and the resulting activation is compared with stored reference assemblies via overlap; the class with highest overlap is predicted.

We adapt this pattern for directional readout. For each ordered pair $(u, v)$, we analytically compute which $k$ neurons in the target area $v$ would win if only the source assembly A\_u were active (plasticity off, deterministic top-$k$ selection):

\[
\texttt{input\_to\_v}[i] = \sum_{j \in A_u} W_{u \to v}[j, i]
\]

\[
\hat{A}_v = \text{top-}k(\texttt{input\_to\_v})
\]

The \textbf{propagation overlap} is the fraction of predicted winners that coincide with the stored target assembly:

\[
\text{overlap}(u \to v) = \frac{|\hat{A}_v \cap A_v|}{k}
\]

Directional evidence follows from the asymmetry of forward and reverse overlaps. For each candidate link $(u, v)$:

\[
\Delta_{\text{prop}}(u, v) = \text{overlap}(u \to v) - \text{overlap}(v \to u)
\]

All ordered pairs are ranked by $\Delta_{\text{prop}}$; Top-K selection and Precision@K / Recall@K evaluation proceed identically to the synaptic-strength readout. This provides a complementary, functionally grounded directional signal: instead of inspecting weight magnitudes, we verify that the trained connectome actually routes activation from source to target assemblies as intended by DIRECT.

\section{Results}\label{results}

The pipeline is evaluated on an Alzheimer SCM (10 variables, 12 directed links; baseline categorical and mixed-input variants) and an education-student-dropout SCM, both over fixed known graphs, with OULAD \cite{kuzilek2017} mappings retained as supplementary portability evidence.

Results are reported as \textbf{evidence for the estimator defined in the Methodology section}, using three complementary check types: single-run structural recovery, fold-partition stability, and perturbation/condition robustness. Across the latest Alzheimer parallel formation-binding runs, the same procedure yields consistent directional recovery under fixed supervision settings.

\begin{itemize}
\tightlist
\item \textbf{Top-K structural recovery}: the Top-K link set matched the ground-truth link set in reported runs.
\item \textbf{Robustness under perturbation}: directional profiles remained stable under source, binding, and stress perturbations (R1--R3).
\item \textbf{Do-calculus validation}: interventional and counterfactual checks confirmed directional consistency using Pearl's backdoor adjustment and SCM-based counterfactual evaluation.
\end{itemize}

The key takeaway is methodological: separating representation stabilization, directed binding, and asymmetry readout produces auditable directionality within ground-truth-guided learning. In the reported runs, the highest-ranked K links (with K fixed to the ground-truth link count) matched the ground-truth link set, yielding complete Top-K recovery under the stated supervision setting.

Evaluation metrics used throughout:

\[
\text{Precision@K} = \frac{\text{TP@K}}{\text{TP@K} + \text{FP@K}}
\]

\[
\text{Recall@K} = \frac{\text{TP@K}}{|\text{ground-truth links}|}
\]

where TP@K and FP@K are the true and false positives among the top-K ranked links.

Here, R1 denotes source-perturbation robustness, R2 denotes binding-perturbation link-localization robustness, and R3 denotes stress-condition robustness.

\begin{itemize}
\tightlist
\item \textbf{R1 (source-perturbation alignment)}: directional preference remained aligned under source perturbation.
\item \textbf{R2 (binding-perturbation localization)}: asymmetry remained link-localized under binding perturbation.
\item \textbf{R3 (stress-condition stability)}: stress conditions did not overturn the base directional profile.
\end{itemize}

To quantify R3 under realistic encoding variation, we ran six random seeds under three encoding-separation conditions (base 15x, milder 10x, harder 6.7x). We vary encoding separation to control representational overlap and test whether causal recovery remains stable as internal encodings become less separable. Structural recovery remained perfect across all 18 runs, while propagation quality stayed high across conditions with bounded variability.

Across all conditions, Precision@K, Recall@K, and propagation pass rate remained 1.0.

\vspace{0.5\baselineskip}
\begin{center}
\small
\begin{tabular}{lcccc}
\hline
\textbf{Condition} & \textbf{Runs} & \textbf{TP} & \textbf{FP} & \textbf{Mean dprop +/- SD} \\
\hline
Base 15x & 6 & \phantom{-}12.0 & \phantom{-}0.0 & 0.801 ± 0.030 \\
Milder 10x & 6 & \phantom{-}12.0 & \phantom{-}0.0 & 0.811 ± 0.026 \\
Harder 6.7x & 6 & \phantom{-}12.0 & \phantom{-}0.0 & 0.787 ± 0.062 \\
\hline
\end{tabular}
\end{center}
\vspace{0.5\baselineskip}

\begin{figure}[H]
\centering
\includegraphics[width=0.92\linewidth]{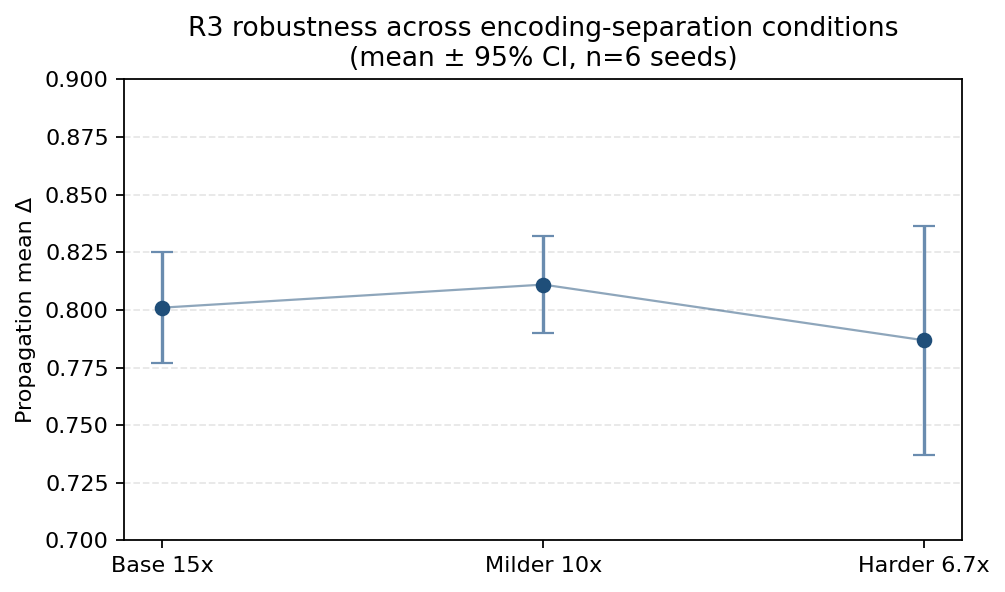}
\caption{R3 robustness summary (mean dprop with 95 percent CI across conditions).}
\end{figure}

Beyond structural recovery and perturbation robustness, we validate directional claims using Pearl's do-calculus \cite{pearl2009}. Do-calculus provides a formal bridge from directional structure to interventional estimands (e.g., $E[Y \mid do(X=x)]$), allowing us to test causal consistency beyond observational association.

\begin{figure}[H]
\centering
\includegraphics[width=0.92\linewidth]{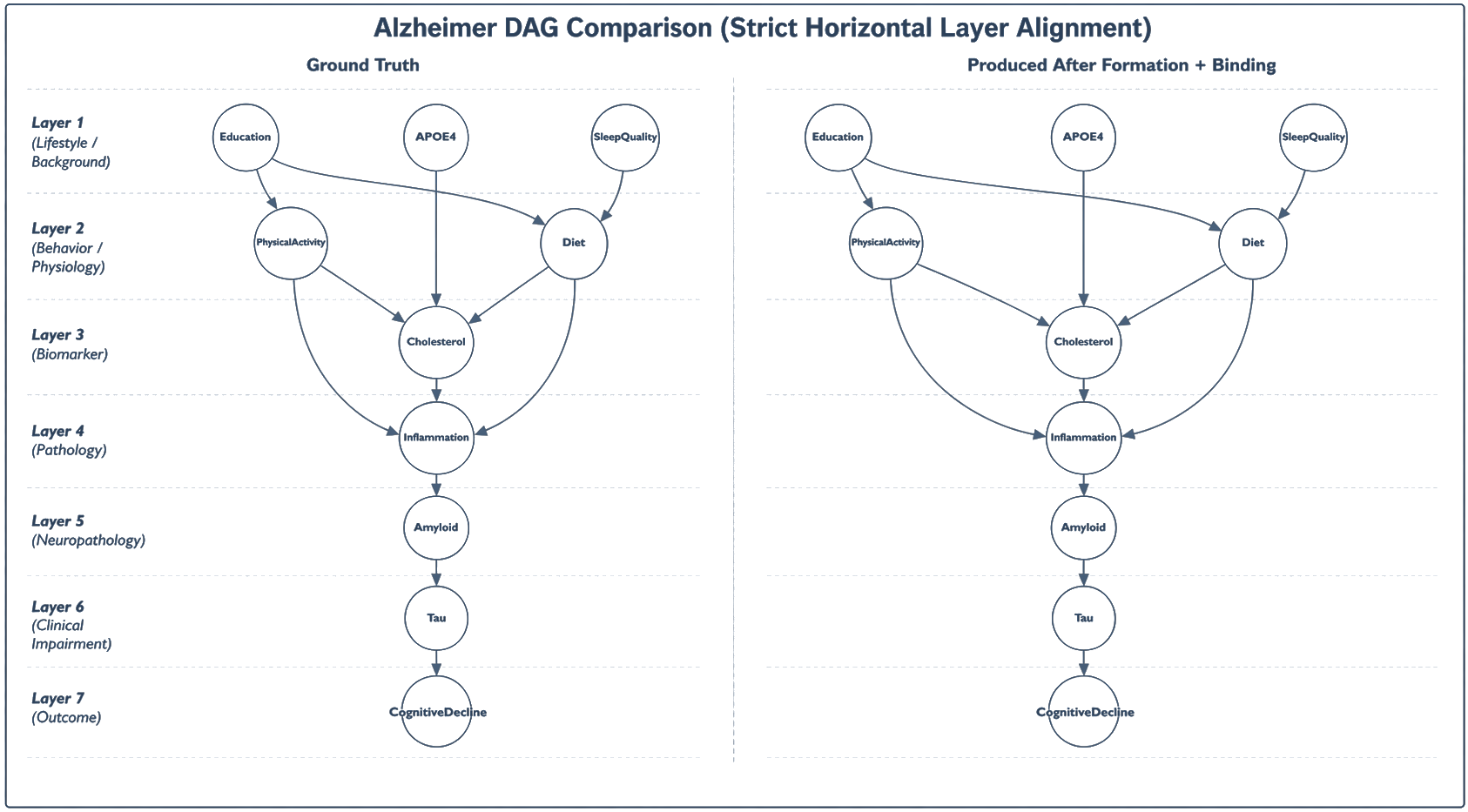}
\caption{Ground-truth vs produced Alzheimer DAG.}
\end{figure}

\textbf{Interventional check.} Interventional effects were validated using covariate adjustment (conditioning on an admissible set Z to block confounding paths). The estimated average treatment effect (ATE) matched the true interventional ATE in sign and magnitude.

\textbf{Counterfactual check.} Counterfactual evaluation (asking what would happen to the same unit under a different treatment) compared factual and counterfactual predictions for the same units under alternative treatment assignments. Direction consistency was confirmed: the learned mechanism preserved the correct sign of the causal effect across counterfactual scenarios generated from the same SCM.

The backdoor adjustment formula underlying the interventional check is:

\[
E[Y \mid do(X = x)] = \sum\limits_{z} E[Y \mid X = x, Z = z] \, P(Z = z)
\]

where \(Z\) is the backdoor adjustment set. Interventional datasets are generated by clamping variables inside the SCM (do-operator), backdoor-adjusted estimates are compared against oracle ATE values, and counterfactual scenarios are evaluated for direction consistency. This makes the causal validation concrete and auditable rather than qualitative.

The propagation overlap readout confirms directional recovery through functional activation flow. On the Alzheimer SCM (10 variables, 12 ground-truth links):

\vspace{0.5\baselineskip}
\begin{center}
\small
\begin{tabular}{lccc}
\hline
\textbf{Pair} & \textbf{fwd overlap} & \textbf{rev overlap} & \(\Delta_{\text{prop}}\) \\
\hline
APOE4\(\to\)Amyloid & \phantom{-}0.740 & \phantom{-}0.130 & \phantom{-}0.610 \\
Education\(\to\)PhysicalActivity & \phantom{-}0.950 & \phantom{-}0.120 & \phantom{-}0.830 \\
Education\(\to\)Diet & \phantom{-}0.990 & \phantom{-}0.040 & \phantom{-}0.950 \\
PhysicalActivity\(\to\)Cholesterol & \phantom{-}1.000 & \phantom{-}0.110 & \phantom{-}0.890 \\
Diet\(\to\)Cholesterol & \phantom{-}1.000 & \phantom{-}0.150 & \phantom{-}0.850 \\
SleepQuality\(\to\)Inflammation & \phantom{-}0.950 & \phantom{-}0.140 & \phantom{-}0.810 \\
Cholesterol\(\to\)Inflammation & \phantom{-}0.950 & \phantom{-}0.120 & \phantom{-}0.830 \\
Inflammation\(\to\)Amyloid & \phantom{-}0.740 & \phantom{-}0.130 & \phantom{-}0.610 \\
Amyloid\(\to\)Tau & \phantom{-}0.990 & \phantom{-}0.040 & \phantom{-}0.950 \\
Tau\(\to\)CognitiveDecline & \phantom{-}1.000 & \phantom{-}0.090 & \phantom{-}0.910 \\
Amyloid\(\to\)CognitiveDecline & \phantom{-}0.980 & \phantom{-}0.070 & \phantom{-}0.910 \\
Education\(\to\)CognitiveDecline & \phantom{-}1.000 & \phantom{-}0.100 & \phantom{-}0.900 \\
\hline
\end{tabular}
\end{center}
\vspace{0.5\baselineskip}

All 12 ground-truth links show positive $\Delta_{\text{prop}}$ (Propagation Precision@K = 1.0, Recall@K = 1.0), matching the synaptic-strength readout. Forward overlaps range from 0.74 to 1.00, while reverse overlaps stay between 0.04 and 0.15, confirming that DIRECT creates functionally directional pathways across the full graph. This provides evidence beyond weight-magnitude inspection---the learned connectome routes information directionally as intended.

\begin{figure}[H]
\centering
\includegraphics[width=0.92\linewidth]{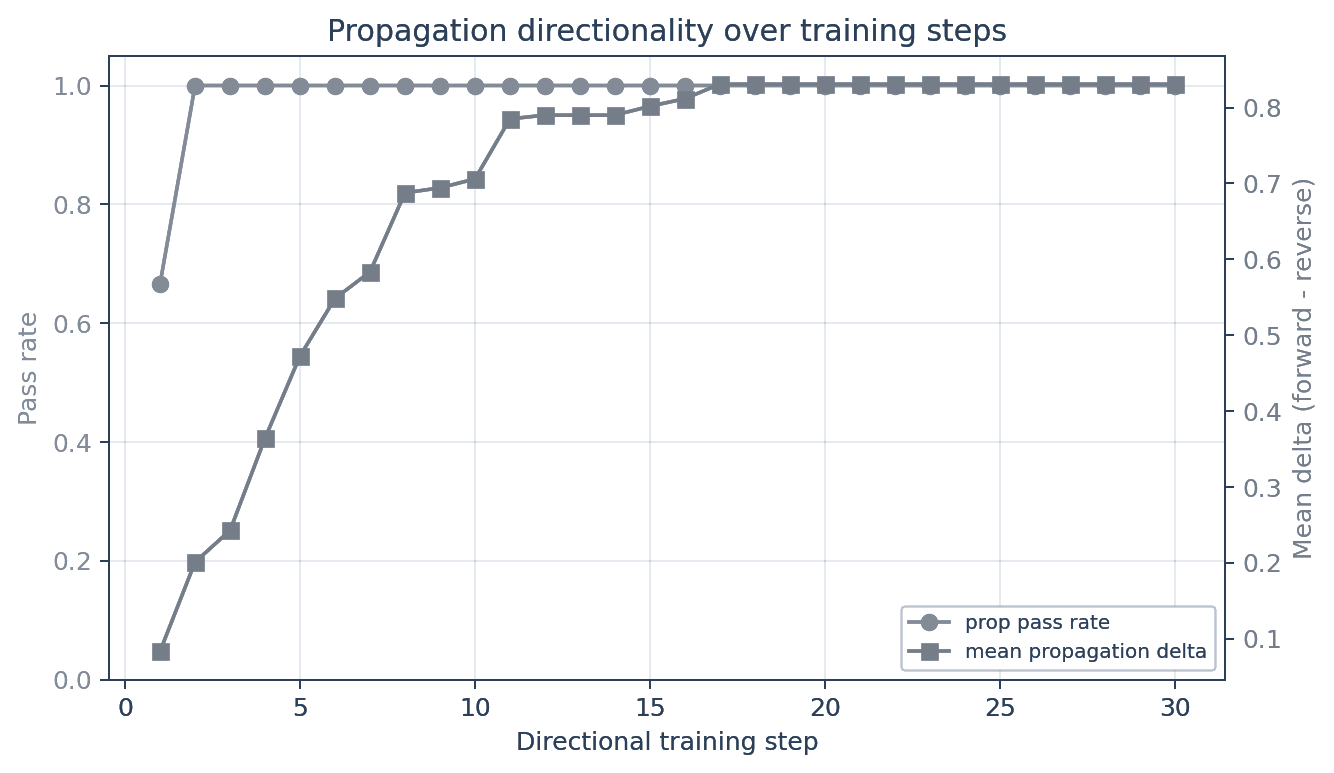}
\caption{Propagation directionality trajectory during supervised binding.}
\end{figure}

\section{Discussion}\label{discussion}

The result profile is strongest for memory replay and bounded rule-following under ground-truth conditions. The mechanism-level transparency is a key strength. Directional behavior is inspectable through neural assembly winners and synaptic asymmetry, not solely through predictive output.

Biological plausibility is preserved because DIRECT does not introduce a global error signal or nonlocal optimizer; it can be viewed as a temporary neuromodulation-like gain bias during paired assembly co-activation during Directional Binding. Synaptic updates remain local, activity-dependent, and winner-sparse (k-capped), with bounded exposures and gain reset after binding episodes. We therefore treat DIRECT as a mechanism-class abstraction compatible with Hebbian/STDP-style plasticity, not as a claim of circuit-faithful biological implementation.

Like prior assembly models in the NEMO line, our model is biologically plausible at the mechanism-class level rather than circuit-faithful; the warm-ramp schedule is an abstract plasticity-control signal, not a detailed biophysical circuit model.

At the same time, we keep strict claim boundaries. This work does not establish observational causal discovery, and it does not claim robust understanding under arbitrary foreign/out-of-distribution (OOD; inputs that differ meaningfully from the training distribution) conditions. Time-aware lagged training is currently treated as an exploratory alternative with limited evidence and should not be interpreted as a standalone causal-direction learner without additional asymmetric directional binding. The validation results therefore support a scope-bounded causal claim, while OOD evidence remains meaningful but partial in breadth. Practically, this should be read as mechanism-valid learning under provided structure and tested perturbation families, not as a universal causal-learning guarantee.

This approach requires externally supplied DAG structure and therefore cannot infer causal direction from observational data alone. Broader real-world transfer across heterogeneous domains remains unproven. The binding-gain schedule is biologically plausible at the mechanism-class level, but it is not intended as a circuit-faithful biological model. OOD coverage is improving but not yet sufficient for broad foreign-data understanding claims.

\section{Conclusions and Further Work}\label{conclusions-and-further-work}

Neural assemblies can be used as a transparent layer for supervised directed connectivity learning when structure is known. The single adaptive schedule (progressively increasing directional coupling over stabilized assemblies) provides interpretable directional parameterization and stable structural recovery under the fixed reporting setup used in the latest Alzheimer parallel formation-binding evaluations. The central value is methodological clarity and mechanism-level auditability under explicit scope control. More broadly, this work operationalizes a missing link between neural assembly computation and formal causal analysis by showing how biologically inspired representation units can be integrated with explicit causal semantics under a reproducible procedure. The strongest contribution is therefore an auditable supervised causal-learning framework with intentionally narrow observational-discovery scope. More broadly, by adding causal directionality to the demonstrated capabilities of neural assemblies — classification, parsing, planning, sequence computation, and statistical learning — this work reinforces the view that assemblies may serve as a computationally general substrate for structured cognition.

Within the ground-truth setting used here, the framework is most suitable for domains where a trusted causal graph is available and mechanism-level auditability is important. Representative use cases include: (i) educational risk modeling with policy-defined dependency structure (as in OULAD-style mappings), (ii) healthcare decision-support workflows where expert-curated variable dependencies are available and directionality must be inspectable, (iii) industrial process monitoring with engineered dependency diagrams, and (iv) scientific simulation pipelines where directional mechanisms are known and stability under setting shifts is required. In all cases, deployment should remain within the explicit scope used in this paper: supervised directionality learning under provided structure, not open-ended observational causal discovery.

This study establishes intervention-backed causal directionality under a ground-truth setting. A natural next step is to expand toward fuller causal learning by integrating structure adaptation under uncertainty, explicit treatment of latent confounding, stronger intervention and counterfactual validation criteria, and transportability/invariance tests across heterogeneous environments. This progression would preserve neural assembly-level transparency while broadening causal guarantees beyond directional consistency under provided structure. A practical near-term priority is to relax supervision assumptions gradually while preserving the same auditability goal.

A concrete next experiment is a temporal-precedence-guided discovery mode that keeps the same assembly mechanism but relaxes full edge supervision. Candidate directed links would be proposed from lagged precedence statistics (for example, how consistently $A$ precedes $B$), then trained with the same adaptive warm-ramp gain policy: conservative directional updates during early winner-set drift, followed by stronger updates after stabilization criteria are met. Evaluation would report edge-level precision/recall, SHD (Structural Hamming Distance; the number of edge additions, deletions, or direction flips needed to match the ground-truth graph), direction accuracy, and robustness across seeds and lag settings. This would test whether the current mechanism can recover useful candidate causal directions under weaker supervision while maintaining explicit limits (temporal precedence as evidence, not a standalone causal-identification guarantee).

An additional future extension would be an agentic AI governance layer that would operate after core biological learning and causal evaluation, not inside them. In this role, an agent could (i) audit whether do-calculus outcomes remain consistent with historical runs and accepted effect ranges, (ii) compare propagation-overlap results against prior checkpoints and alternative configurations, and (iii) prioritize follow-up experiments when consistency degrades. This would keep transition decisions biologically grounded while potentially improving reliability, traceability, and experimental planning.

\bibliographystyle{ACM-Reference-Format}
\bibliography{references}

\end{document}